\UseRawInputEncoding
\documentclass[sigconf]{acmart}
\pdfoutput=1
\renewcommand\footnotetextcopyrightpermission[1]{} 

\AtBeginDocument{%
	\providecommand\BibTeX{{%
			\normalfont B\kern-0.5em{\scshape i\kern-0.25em b}\kern-0.8em\TeX}}}

\usepackage{amsmath}
\usepackage{graphicx}
\usepackage{amsthm,amsmath,amsfonts}
\usepackage{subfigure}
\usepackage{color}
\usepackage{bm}
\usepackage{multirow}

\def\bA{\textbf{A}}

\def\bH{\textbf{H}}

\def\bh{\textbf{h}}
\def\bz{\textbf{z}}
\def\bal{\bm{\alpha}}
\usepackage[normalem]{ulem}
\useunder{\uline}{\ul}{}

\usepackage{colortbl}

\newcolumntype{L}[1]{>{\raggedright\let\newline\\\arraybackslash\hspace{0pt}}m{#1}}
\newcolumntype{C}[1]{>{\centering\let\newline  \\\arraybackslash\hspace{0pt}}m{#1}}
\newcolumntype{R}[1]{>{\raggedleft\let\newline \\\arraybackslash\hspace{0pt}}m{#1}}
\begin{document}
\title{Enhancing Intra-class Information Extraction for Heterophilous Graphs: One Neural Architecture Search Approach}

\author{Lanning Wei$^{1,2,3}$, 
		Zhiqiang He$^{1,4}$,
	Huan Zhao$^3$, 
	Quanming Yao$^{5}$}
\affiliation{
	\institution{$^1$Institute of Computing Technology, Chinese Academy of Sciences $^2$University of Chinese Academy of Sciences $^3$4Paradigm. Inc.,$^4$Lenovo, $^5$Department of Electronic Engineering, Tsinghua University}
	\city{Beijing}
	\country{China}
}
\email{weilanning18z@ict.ac.cn; hezq@lenovo.com; zhaohuan@4paradigm.com; qyaoaa@tsinghua.edu.cn}

\begin{abstract}
	In recent years, Graph Neural Networks (GNNs) have been popular in graph representation learning which assumes the homophily property, i.e., the connected nodes have the same label or have similar features. However, they may fail to generalize into the heterophilous graphs which in the low/medium level of homophily. 
	Existing methods tend to address this problem by enhancing the intra-class information extraction, i.e., either by designing better GNNs to improve the model effectiveness, or re-designing the graph structures to incorporate more potential intra-class nodes from distant hops. 
	Despite the success, we observe two aspects that can be further improved: (a) enhancing the ego feature information extraction from node itself which is more reliable in extracting the intra-class information; (b) designing node-wise GNNs can better adapt to the nodes with different homophily ratios.
	In this paper, we propose a novel method IIE-GNN (Intra-class Information Enhanced Graph Neural Networks) to achieve two improvements. A unified framework is proposed based on the literature, in which the intra-class information from the node itself and neighbors can be extracted based on seven carefully designed blocks. With the help of neural architecture search (NAS), we propose a novel search space based on the framework, and then provide an architecture predictor to design GNNs for each node.
	We further conduct experiments to show that IIE-GNN can improve the model performance by designing node-wise GNNs to enhance intra-class information extraction.
\footnote{Lanning and Zhiqiang contribute equally to this work, and Lanning is a research intern in 4Paradigm. Huan Zhao and Quanming Yao are the corresponding authors. }

\end{abstract}

\maketitle
\section{Introduction}
Graph Neural Networks (GNNs) have been widely used in recent years due to their promising performance on diverse graph-based applications~\cite{kipf2016semi,ying2018hierarchical}. In the literature, most of them are designed based on the message-passing scheme~\cite{gilmer2017neural}, i.e, aggregating the messages from the connected neighbors. These methods achieve great success on real-world datasets constructed under homophily~\cite{zhu2020beyond}, i.e., the connected nodes belong to the same class or have similar features. On the contrary, for those heterophilous graphs in the opposite situation, GNNs get unpredictable performance degradations and are even outperformed by MLPs (Multi-layer perceptron).  The features from different classes are mixed directly and lead to indistinguishable node representations in the node classification task~\cite{zhu2020beyond}. 

When facing heterophilous datasets, how to designing expressive GNNs is the key challenge. Several methods tend to enlarge the receptive field by incorporating more intra-class nodes. For example, \cite{pei2020geom,suresh2021breaking} incorporates the distant nodes which may exhibit similar structure and in the same class; \cite{zhu2020beyond,jin2021universal} utilize the higher-order neighbors which may contain more nodes with the same label. Apart from these methods that use non-local neighbors, more methods tend to refine the GNN architectures.  \cite{zhu2020beyond,hamilton2017inductive} separate the ego and neighbor information in the GNN layer which enables the evolving of the non-aggregated features; 
\cite{du2022gbk,yan2021two,bo2021beyond} aim to learn edge weights to separate the homophilous and heterophilous information, e.g., attaching the larger weights to homophilous neighbors while the smaller or even negative weights for the heterophilous ones.
Utilizing the inter-layer connections is also popular in designing GNNs for heterophilous graphs~\cite{zhu2020beyond,chien2020adaptive} which claims to incorporate the high-frequency components in the shallow layers.

Existing methods aim to enhance the intra-class information, either by increasing the number of intra-class nodes in the receptive fiel, or enlarging the edge weights from the intra-nodes. 
In contrast, the ego-feature acts as a critical element in the intra-class information, which is much more reliable compared with the intra-class information collected from neighbor features. However, existing methods only consider the ego features when (a) updating the node features by separating the ego and neighbor's features; (b) integrating the features from the intermediate layers.
It is insufficient to enhance the ego features compared with the widely designed methods, and it is challenging to design an effective and comprehensive way to enhance the intra-class information extraction from two parts, i.e., ego nodes and other intra-class nodes.

\begin{figure}
	\centering
	\includegraphics[width=0.8\linewidth]{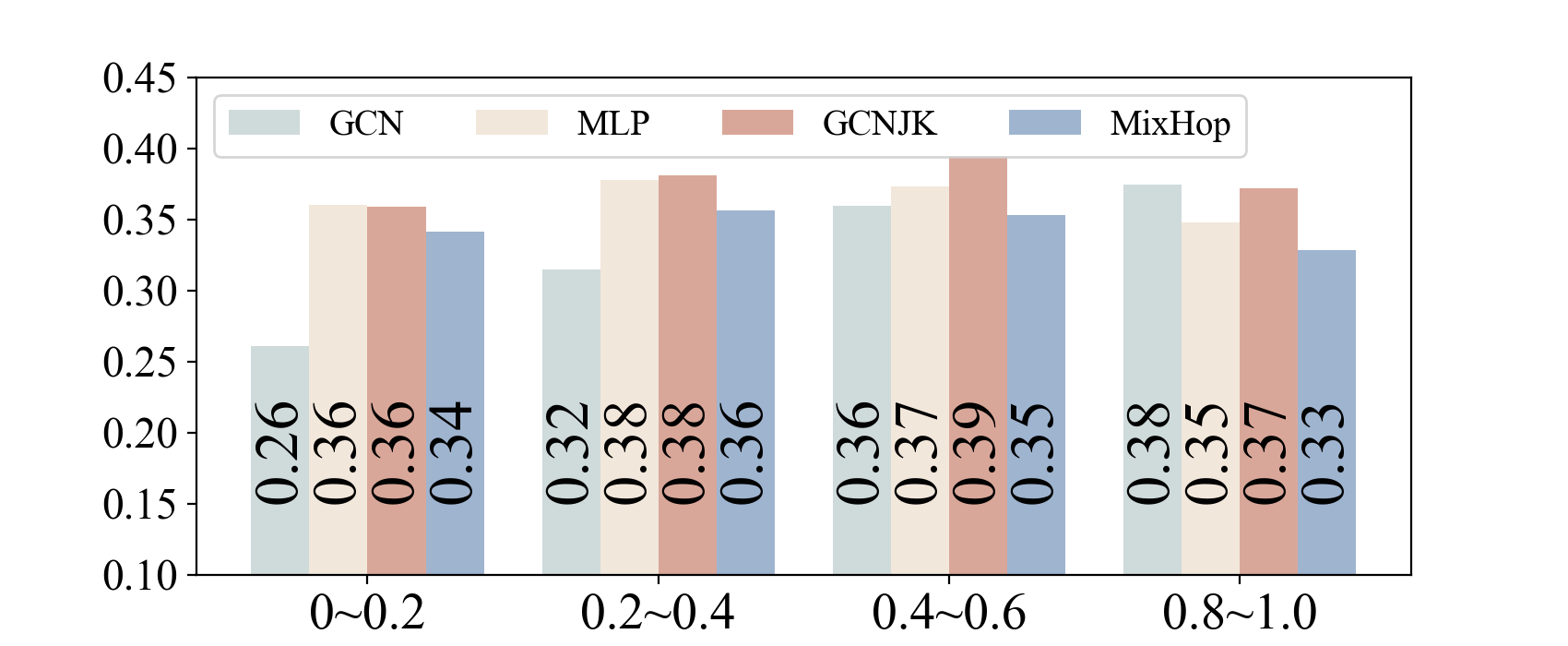}
	\caption{The illustrations of diffferent baselines on nodes with homophilous ratios in Actor dataset.}
	\label{fig-hacc}
\end{figure}

Besides, reviewing the heterophilous graphs from the node-level,  they have nodes with diverse homophily ratios as shown in Figure~\ref{fig-hacc}, on top of which different baselines have diverse performance on these nodes. Therefore, existing methods, which are designed to be shared among all nodes in the graph, are inrational to be applied to these various nodes considering the diverse node homophily ratios~\cite{wang2022graph,du2022gbk}.
However, it is extremely time-consuming to achieve this with human expertise. How to designing node-wise GNNs is a challenging problem, especially considering large-scale graphs.

In this paper, we propose a novel method to address the aforementioned challenges: (a) designing a unified framework based on the literature, which could incorporate the comprehensive intra-class information from the node itself and neighbors in different levels; (b) with the help of neural architecture search (NAS), we design a novel search space based on the unified framework, and then providing an architecture predictor to seleting architectures.

To be specific, seven blocks in four levels are derived from the literature: 
selection and attention blocks in the neighbor-level responsible for selecting and reweighing potential neighbors (besides node itself) which could provide the intra-class information; 
aggregation and update blocks in the message-level aim to constructing the messages and updating the node features based on these collected potentional intra-class feature vectors. These four blocks make up one gnn layer. 
Then, the residual-merge and inter-merge blocks in the layer-level are provided to incorporating the features of node itself from different layers to imporve the model effectiveness; then the output-merge block is provided to incorporate the MLP branch which generates the non-aggregated node features.
Based on this framework, we construct an effective search space by providing a set of candidate operations in each block, e.g., selecting whether to use the ego- or neighbor information merely or mixing them. 
When designing the node-wise GNNs, we adopt the architecture predictor to calculate the node-aware operation distributions which could separate the influence of node numbers. 
Considering the largely different features extracted from the node itself and its neighbors~\cite{zhu2020beyond},
we design an architecture predictor which firstly concatenate all the inputs, on top of which the operation distributions can be generated with the help of  architecture predictors.
We conduct extensive experiments on heterophilous datasets, and the proposed method could achieve the SOTA performance by enhancing the intra-class information extraction with carefully designed node-wise GNNs.

%


To summarize, the contributions of this paper are as follows:

\begin{itemize}
	\item We propose a novel method IIE-GNN to design the self-enhanced graph node-wise architectures for heterophilous graphs.
	To effectively use the ego features, we design an effective framework which contains seven blocks in four levels, and it can unify the literature and further enrich the intra-class information.
	\item  To design the node-aware GNNs, we develop a set of operations in each block to construct the supernet on top of the proposed framework. Two architecture predictors are designed to select node-wise architectures considering the efficiency and the heterophilous nature.
	\item Extensive experiments are conducted on five heterophilous graphs, and the SOTA performance demonstrate the effectiveness of the proposed IIE-GNN.
\end{itemize}

\section{Related Work}
\subsection{Preliminary}
\label{sec-preliminary}
\noindent\textbf{Notations.} 
For a graph $\mathcal{G} =(\mathcal{V}, \mathcal{E}) $,  $\mathcal{V}$ and $\mathcal{E}$ represent the node and edge sets in this graph. $\textbf{A} \in \mathbb{R}^{|\mathcal{V}| \times |\mathcal{V}|}$ is the adjacency matrix of this graph where $|\mathcal{V}|$ is the node number, and $\mathcal{N}(u), \tilde{\mathcal{N}}(u)$ and $\bar{N}(u)$ represent the connected neighbors without self-loop, neighbors with self-loop and the selected neighbors for node $u$. The class of node $u$ is represented as $y_u$.

\noindent\textbf{Homophily ratio.} There are two metrices to evaluate the homophily ratio: the edge homophily ratio $h_{edge}$ and the node homophily ratio $h_{node}$ as shown in the following:
\begin{align}
	\nonumber
	h_{edge} = \frac{\left| \{(u,v):(u,v) \in \mathcal{E} \wedge y_u=y_v\}\right|}{\left| \mathcal{E} \right|}, \\ 
	\nonumber
	h_{node} = \frac{1}{|\mathcal{V}|} \sum_{v \in \mathcal{V}} \frac{\{u| u \in \mathcal{N}(v) \wedge y_u=y_v\}}{d_v}. 
\end{align}

\subsection{GNNs designed under heterophily}
Observing that GNNs which designed under homophily perform poorly on those heterophilous graphs~\cite{pei2020geom,zhu2020beyond}, researches tend to design GNNs for those heterophilous graphs and they can be categorized into two categories~\cite{zheng2022graph}, i.e., (a) incorporating the non-local neighbors which may have the same label~\cite{pei2020geom,suresh2021breaking,jin2021universal,wang2022graph}; 
and (b) refining the GNNs architecture by re-designing the GNN layers~\cite{du2022gbk,yan2021two,fang2022polarized,zhu2020beyond} or designing the inter-layer connections and layer weightes~\cite{wei2022designing,zhu2020beyond,chen2020measuring}.  More methods can be found in~\cite{zheng2022graph}.

\subsection{Graph Neural Architecture Search}
Very recently, researchers tried to automatically design GNN architectures by NAS. The majority of these methods focus on designing the aggregation layers under homophily setting.
For example, \cite{gao2019graphnas,zhou2019auto,yoon2020autonomous,li2021one} learn to design aggregation layers on top of several micro dimensions, such as aggregation function, attention function, attention head number, embedding size, etc; \cite{zhao2021search,zhao2020simplifying,li2020autograph,wei2022designing} provide the extra layer-wise skip connections design dimension; \cite{cai2021rethinking,lai2020policy,wang2022graph} learn to design the depth of GNNs. More methods can be found in \cite{zhang2021automated}.
%

\section{Method}

To enhance the intra-class information extraction when designing GNNs on the heterophilous graphs, we first provide a unified framework which split GNNs into seven blocks into four levels, on top of which we design a novel search space and a architecture predictor to design node-wide GNNs.

%
%
%

\begin{figure*}
	\centering
	\includegraphics[width=0.95\linewidth]{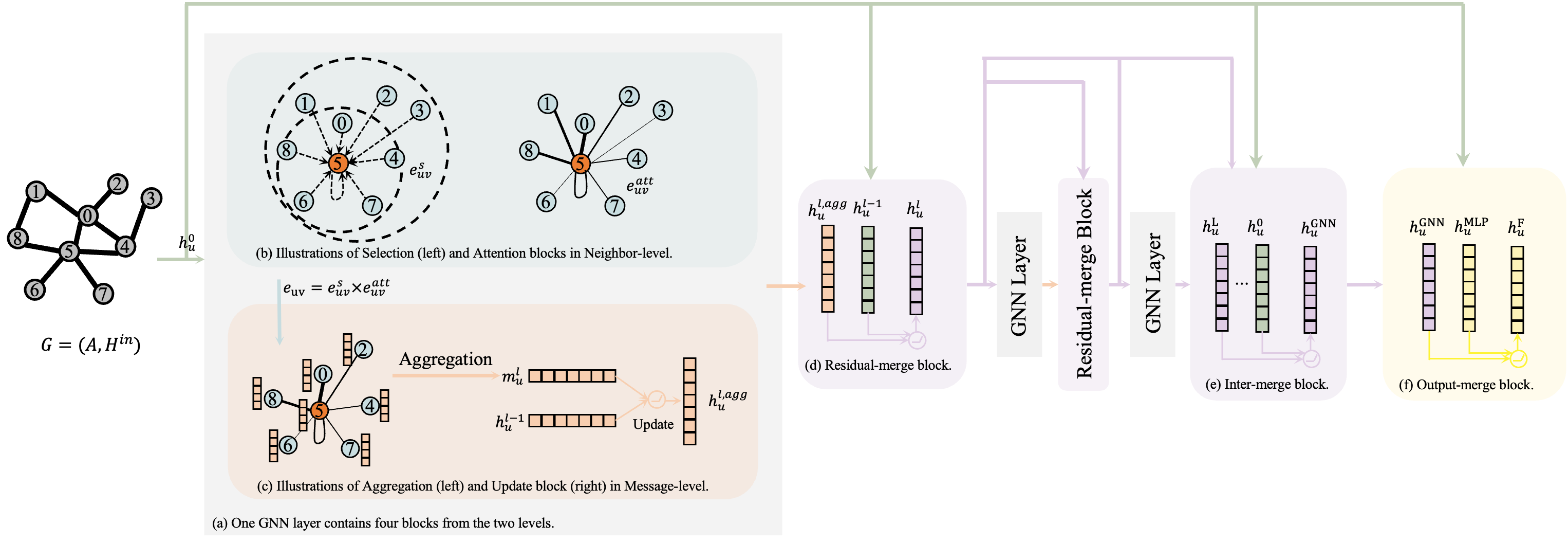}
	\caption{An overview of the proposed framework. (a) Each GNN layer contains two blocks in the Neighbor-level (as shown in (b)) and two blocks in the Message-level (as shown in (c)). (d-e) The designed Residual-merge and Inter-layer blocks in the Layer-level. (f) The designed output-merge block in the Network-level.
	}
	\label{fig-framework}
\end{figure*}

\subsection{The Unified Framework to Extract the Intra-class Information}	
General GNNs focus on designing the aggregation operation in each layer and the network topology with the inter-layer connections. The aggregation operation is designed under the message-passing scheme~\cite{gilmer2017neural}. 
Considering the heterophilous graphs in which the node and neighbors are from different classes in general, utilizing the intra-class information and enhancing the ego feature extraction are significant~\cite{zhu2020beyond,bo2021beyond,suresh2021breaking}. 
To achieve this, we defined a novel framework as shown in Figure~\ref{fig-framework}. 
It contains four levels with seven blocks to designing the GNNs for those heterophilous graphs.

In each GNN layer, we provide the selection and attention blocks to choose and re-weight the neighbors.  The former is responsible for choosing neighbors $\bar{\mathcal{N}}(u)$ which is used to construct the messages, and it is expected to incorporate more potential intra-class nodes.
In this paper, the selection block is achieved by designing the edge weights $e_{uv}^s =  \{1| \forall v \in \bar{\mathcal{N}}(u)\}$.
The attention block aims to re-weight the neighbors, i.e., designing the edge weights $e_{uv}^{att}$ for each potential neighbor
With the help of these two blocks, more potential intra-class neighbors can be incorporated and emphasized by attaching different edge weights $e_{uv} = e_{uv}^s\times e_{uv}^{att}$, after which the intra-class information are provided.
After that, we provide a aggregation block and a update block in the message level to integrate the features from different neighbors into the node embeddings, which are also indispensable stages in the message-passing scheme.
The aggregation block aims to construct the message based on the collected weighted neighbor features.
The update block devotes to updating the node embeddings based on the messages and the node feature itself, which are denoted as $m_u^l$ and $\bh_u^{l-1}$ as shown in Figure~\ref{fig-framework}(c).

Beyond the GNN layer, we provide the residual-merge and the inter-merge blocks to improve the model's effectiveness by utilizing the ego features in different layers. In this paper, the former block considers merging the ego features from the previous consecutive layers, i.e., enhancing the feature utilization based on the aggregation results $\bh_u^{l,agg}$ and the residual vector $\bh_u^l$.
The inter-merge block is employed at the end of GNN layers, and it aims to merge the ego features from different layers, i.e., $\{\bh_u^l|0\leq l \leq L\}$ in the $L$-layer GNNs.Beyond GNN, MLP is the comparable network for the heterophilous graphs~\cite{ma2021homophily}, which can make up the ego information that may be lost in the aggregation operations. To utilize these features, we use the two-layer MLP to generate the non-aggregated features and provide a output-merge block before the classifier. 

\subsection{Designing Search Space Based on the Framework.}
\label{sec-search-space}
As shown in Table~\ref{tb-search-space}, we provide a set of candidate operations in each block to construct the search space.

$\bullet$ Selection block. We provide two operations \texttt{1N} and \texttt{1LOOPN} to select different neighbors. The former operation only uses the one-hop neighbors $\bar{\mathcal{N}}(u) = \mathcal{N}(u)$ and sets the edge weights into 1. The latter uses $\bar{\mathcal{N}}(u) = \tilde{\mathcal{N}}(u)$ which further incorporate the self-loop edges.

$\bullet$ Attention block. Four attention functions are derived from the widely used methods.
\texttt{CONST} attaches the same weights for each neighbor, i.e., $e_{uv}^{att} = 1$, and \texttt{SYM\_NORM}~\cite{kipf2016semi} uses the symmetric normalized adjacency;  \texttt{GATE\_FILTER}~\cite{bo2021beyond} and \texttt{SIGNED}~\cite{yan2021two} are designed for the heterophilous graphs especially, which are expected to attach large weights to the intra-class neighbors while the small or even negative weights for inter-class ones.

$\bullet$ Aggregation block. We provide a operation \texttt{ADD} which adds up all the features to construct the messages, and it can be denoted as $m_u = \sum _{v \in \bar{\mathcal{N}}(u)}{e_{uv}\bh_v}$. Other aggregation operations utilized bi-kernel for different neighbors~\cite{du2022gbk} can also be utilized as well. 



$\bullet$ Inter-merge block. Four operations in this block merge ego features from all the intermediate layers: \texttt{SUM} and \texttt{MEAN} represent merging these features with summation and average, respectively. \texttt{LEARN\_ATT} is extracted from GPRGNN~\cite{chien2020adaptive} which learns the layer weights adaptively, and the signed values are available. \texttt{NONSKIP} represent GNN only adopted $\bH^L$ and dropout the features from the intermediate layers.


$\bullet$ Update / Residual-merge / Output-merge block. For these three blocks, two input features are provided which contain intra-class information from the node itself and the generated based on the neighbors, i.e., $(\bh_u^{l-1}, m_u^l)$, $(\bh_u^{l-1}, \bh_u^{l,agg})$, and $(\bh_u^{MLP}, \bh_u^{GNN})$ in these three blocks, respectively.
we provide five operations to utilize the intra-class information among them: \texttt{EGO} / \texttt{RES} / \texttt{MLP} in three blocks utilize the ego features merely, while \texttt{MSG} / \texttt{AGG} / \texttt{GNN} preserve the intra-class information which generated from the neighbors. Apart from that, \texttt{SUM}, \texttt{MEAN} and \texttt{ATT} operations are provided additionally which represent these features are merged with the summation, average and attention mechanism.


\begin{table}[]
	\footnotesize
	\caption{The operations used in the search space. }
\begin{tabular}{c|c|c}
	\hline
	Level & Block                          & Operations                                                                 \\ \hline
	\multirow{2}{*}{Neighbor} & Selection $\mathcal{O}_{se}$       & \texttt{1N}, \texttt{1LOOPN}                                               \\ \cline{2-3} 
	& Attention $\mathcal{O}_{att}$      & \texttt{CONST}, \texttt{SYM\_NORM}, \texttt{GATE\_FILTER}, \texttt{SIGNED} \\ \hline
	\multirow{2}{*}{Message}  
	& Aggregation $\mathcal{O}_{aggr}$   & \texttt{ADD}                                                               \\ \cline{2-3} 
	& Update $\mathcal{O}_{update}$      & \texttt{EGO}, \texttt{NEIGHBOR}, \texttt{SUM}, \texttt{MEAN}, \texttt{ATT} \\ \hline
	\multirow{2}{*}{Layer}    
	& Residual-merge  $\mathcal{O}_{rm}$ & \texttt{RES}, \texttt{AGG}, \texttt{SUM}, \texttt{MEAN}, \texttt{ATT}      \\ \cline{2-3} 
	& Inter-merge $\mathcal{O}_{im}$                       & \texttt{SUM}, \texttt{MEAN}, \texttt{LEARN\_ATT}, \texttt{NONSKIP}         \\ \hline
	Network                   & Output-merge $\mathcal{O}_{om}$                      & \texttt{MLP}, \texttt{GNN}, \texttt{SUM}, \texttt{MEAN}, \texttt{ATT}      \\ \hline
\end{tabular}
\label{tb-search-space}
\end{table}

\subsection{Node-aware GNN through Architecture Predictor}



\subsubsection{The inefficiency of vanilla differentiable node-wise GNN design.}
Considering the search efficiency, the differentiable search algorithm is widely used in this paper~\cite{xie2018snas,liu2018darts}. When using these methods to design GNNs, a learnable parameter $\alpha$ is required for each candidate operation and, and then the mixed results can be calculated in each block, i.e.,  $\bar{o}(x) = \sum_{i=1}^{|\mathcal{O}|}c_io_i(x)$ where $c_i$ is the operation weights generated based on the corresponding parameter $\alpha_i$. 
However, when designing node-wise GNNs, a learnable parameter is required for each candidate operation in each node, 
e.g., the learnable selection operations for all nodes can be denoted as $\bal_{se} \in \mathbb{R}^{N \times |\mathcal{O}_{se}|  \times L}$. 
In this situation, the learnable parameter number is proportional to the node number, which increases the optimization difficulty for those large-scale graphs. Therefore, it is inefficient to employ the general differentiable architecture search methods when designing the node-wise GNNs.

%

\subsubsection{The deficient of vanilla architecture predictor.}
One straightforward method is designing an architecture predictor which decouples the node numbers and the learnable parameters. As shown in the following:
\begin{align}
	\label{eq-vanilla-predictor}
 c_i = \frac{\text{exp}(\alpha_i / \tau)}{\sum_{j=1}^{|\mathcal{O}|} \text{exp}(\alpha_j / \tau)},
	\bm{\alpha} = f_{\alpha}(x), f_{\alpha}: \mathbb{R}^{d} \rightarrow \mathbb{R}^{|\mathcal{O}|},
\end{align}
$x$ is the node feature vector, $f_{\alpha}$ is the architecture predictor which mapping the $d$-dim node features into the $|\mathcal{O}|$-dim operation weights $\bm{\alpha}$, and $\alpha_i$ is the $i$-th dimension. In general, MLPs can be employed as the architecture predictor for each block. Then the parameter number is correlated with this neural network design, i.e., the layer number and hidden size,  rather than the node numbers.

However, it is unsuitable to be directly applied to the proposed search space mentioned in Section~\ref{sec-search-space} due to the candidate operations which merely preserve the intra-class features either from the node itself or neighbors.
For the update, residual-merge, inter-merge and output-merge blocks, they will receive two input features that are different from each other considering the heterophilous properties~\cite{zhu2020beyond}.
Then, designing the operations weights based on either of them is not suitable, e.g., calculating operation weight for \texttt{NEIGHBOR} based on  $\bal = f_{\alpha}(\bh_u^{l-1})$.

\subsubsection{Designing the inputs of architectures predictors.}
To avoid the operations weights designed on the partial inputs, we first concatenates all the inputs and then design the operation weights.
Use the update block as an example, 
\begin{align}
		\bm{\alpha}=f_{\alpha}(h_u^{l-1}||m_u^l), f_{\alpha}: \mathbb{R}^{2d} \rightarrow \mathbb{R}^{|\mathcal{O}|}, 
\end{align}
$d$ is the dimension of each input vector, and the predictor $f_{\alpha}$ mapping the cancatenated features into the operations weights $\bal$.

Then, our method can be optimized with the gradient descent which introduced in~\cite{liu2018darts,xie2018snas}. After finishing the search process, we preserve the operation with the largest weight in each mixed operation and each node, from which we obtain the searched architectures.

\section{Experiments}
\subsection{Experimental settings}

\begin{table}[]
	\caption{Statistics of the eight datasets in our experiments.}
	\label{tb-statistic}
	\begin{tabular}{c|c|c|c|c}
		\hline
		Datasets       & \# Nodes  & \# Edges   & \#Features  & \# Classes\\ \hline
		Squirrel   & 5,201   & 198,493   & 2,089  & 5 \\ \hline
		Chameleon  & 2,277   & 31,421    & 2,325  & 5 \\ \hline
		Actor      & 7,600   & 30,019    & 932    & 5 \\ \hline
		Texas      & 183     & 309       & 1,703  & 5 \\ \hline
		Arxiv-year & 169,343 & 1,166,243 & 128    & 5 \\ \hline
	\end{tabular}
\end{table}

In this paper, we adopt five heterophilous graphs to evaluate the proposed IIE-GNN.
Three kinds of baselines are utilized in this paper. 
(a) General GNNs designed under homophily: three-layer GCN~\cite{kipf2016semi}, GraphSAGE~\cite{hamilton2017inductive}, MixHop~\cite{abu2019mixhop}, and the method GCN-JK which mergr the JK-Net~\cite{xu2018representation} and GCN methods;
(b)  GNNs designed suitable for heterophilous graphs: Geom-GCN~\cite{pei2020geom} and WRGNN~\cite{suresh2021breaking} extend the non-local neighbors into the aggregation operations; H2GCN~\cite{zhu2020beyond}, GBK-GNN~\cite{du2022gbk} and GGCN~\cite{yan2021two} refine the GNNs to improve the representation power. Besides, we also adopt the results mentioned in~\cite{ma2021homophily}, which is denoted as MLP+GCN in this paper.
(c) The NAS-based methods suitable for heterophilous graphs: F2GNN~\cite{wei2022designing} aims to design the network topology, and NW-GNN~\cite{wang2022graph} aims to design node-wise GNNs for both homophilous and heterophilous graphs.

We adopt nine heterophilous graphs in this paper to evaluate the proposed IIE-GNN, and the statistics of these datasets are provided in Table~\ref{tb-statistic}. 
For arxiv-year datasets, we adopt the 5 random splits used in~\cite{lim2021new} which has 50\%/25\%/25\% or nodes per class for train/validation/test, respectively. For the other four widely-used heterophilous datasets proposed in ~\cite{pei2020geom,zhu2020beyond}, we adopt the 10 random splits which are the same as~\cite{pei2020geom} (48\%/32\%/20\% of nodes per class).
For our method, we first design GNNs based on the designed search space. all the searched GNNs and the human-designed baselines are tuned individually with hyperpa-rameters like embedding size, learning rate, dropout, etc. With the searched hyperparameters, we report the average test accuracy and the standard deviation as shown in Table~\ref{tb-performance}.

\begin{table*}[]
	\caption{Left: Performance comparisons of our method and all baselines. We report the average test accuracy and the standard deviation with 10 splits. For the methods marked with ``*'', the results are obtained from the corresponding paper. ``NA'' represented the results are not reported in the paper.  The best result in each dataset is highlighted in gray, and the second best one is underlined. Right: The operation distributions in each block in the Texas dataset. ``L1$\_{\mathcal{O}_{se}}$'' represent the selection block in the first GNN layer, and others are the same.}
	\label{tb-performance}
\begin{minipage}{0.7\linewidth}
\footnotesize
\begin{tabular}{c|c|c|c|c|c|c}
\hline
 &  & Squirrel & Chameleon & Actor & Texas & Arxiv-year \\ \hline
& h & 0.22 & 0.23 & 0.22 & 0.11 &0.27  \\ \hline
& MLP & 0.3168(0.0190) & 0.4811(0.0223) & 0.3617(0.0109) & 0.8330(0.0454) & 0.3661(0.0012) \\ 
& GCN~\cite{kipf2016semi} & 0.4974(0.0256) & 0.6728(0.0211) & 0.2816(0.0155) & 0.5757(0.0580) & 0.4675(0.0019) \\ 
& MixHop~\cite{abu2019mixhop} & 0.3607(0.0151) & 0.4945(0.0273) & 0.3301(0.0074) & 0.7135(0.0757) & 0.4829(0.0026) \\ 
& GraphSAGE~\cite{hamilton2017inductive} & 0.3764(0.0240) & 0.5568(0.0236) & 0.3461(0.0093) & 0.7378(0.0746) & 0.4883(0.0008) \\ 
\multirow{-5}{*}{Baseline} & GCN-JK~\cite{xu2018representation} & 0.3784(0.0181) & 0.6167(0.0173) & 0.3616(0.0161) & 0.8027(0.0528) & 0.4652(0.0017) \\ \hline
& Geom-GCN*~\cite{pei2020geom} & 0.3814 & 0.609 & 0.3163 & 0.6757 & NA \\ 
& UGCN*~\cite{jin2021universal} & 0.3439 & 0.5407 & NA & 0.7172 & NA \\ 
\multirow{-3}{*}{\begin{tabular}[c]{@{}c@{}}Extend\\ non-local \\ GNNs\end{tabular}} & \cellcolor[HTML]{FFFFFF}WRGNN*~\cite{suresh2021breaking} & 0.4885(0.0078) & 0.6524(0.0087) & 0.3653(0.0078) & 0.8362(0.0550) & NA \\ \hline
& MLP+GCN*~\cite{ma2021homophily} & 0.5448(0.0111) & 0.6804(0.0186) & 0.3624(0.0109) & 0.8360(0.0654) & NA \\ 
& H2GCN*~\cite{zhu2020beyond} & 0.3642(0.0189) & 0.5711(0.0158) & 0.3586(0.0103) & 0.8486(0.0677) & {\ul 0.4909(0.0010)} \\ 
& GBKGNN*~\cite{du2022gbk} & 0.5590(0.0114) & 0.6159(0.0221) & 0.4289(0.0097) & 0.8061(0.0488) & NA \\ 
\multirow{-4}{*}{\begin{tabular}[c]{@{}c@{}}Refine \\ GNNs\end{tabular}} 
& GGCN*~\cite{yan2021two} & 0.5517(0.0158) & {\ul 0.7114(0.0184)} & 0.3754(0.0156) & 0.8486(0.0455) & NA \\ \hline
& NW-GNN*~\cite{wang2022graph} & {\ul 0.5664(0.0048)} & 0.6906(0.0093) & \cellcolor[HTML]{C0C0C0}0.4448(0.0069) & 0.8180(0.0359) & NA \\ 
\multirow{-2}{*}{NAS-GNNs} 
& F2GCN*~\cite{wei2022designing} & 0.4181(0.0157) & 0.6331(0.0086) & 0.3701(0.0101) & 0.8378(0.0527) & NA \\ \hline
\multicolumn{2}{c|}{IIE-GNN} & \cellcolor[HTML]{C0C0C0}0.5732(0.0189) & \cellcolor[HTML]{C0C0C0}0.7213(0.0211) & {\ul 0.3991(0.0241)} & \cellcolor[HTML]{C0C0C0}{\ul 0.8584(0.0423)} & \cellcolor[HTML]{C0C0C0}0.5159(0.0024) \\ \hline
\end{tabular}
\end{minipage}
\hfill
\begin{minipage}{0.25\linewidth}
	\includegraphics[width=0.9\linewidth]{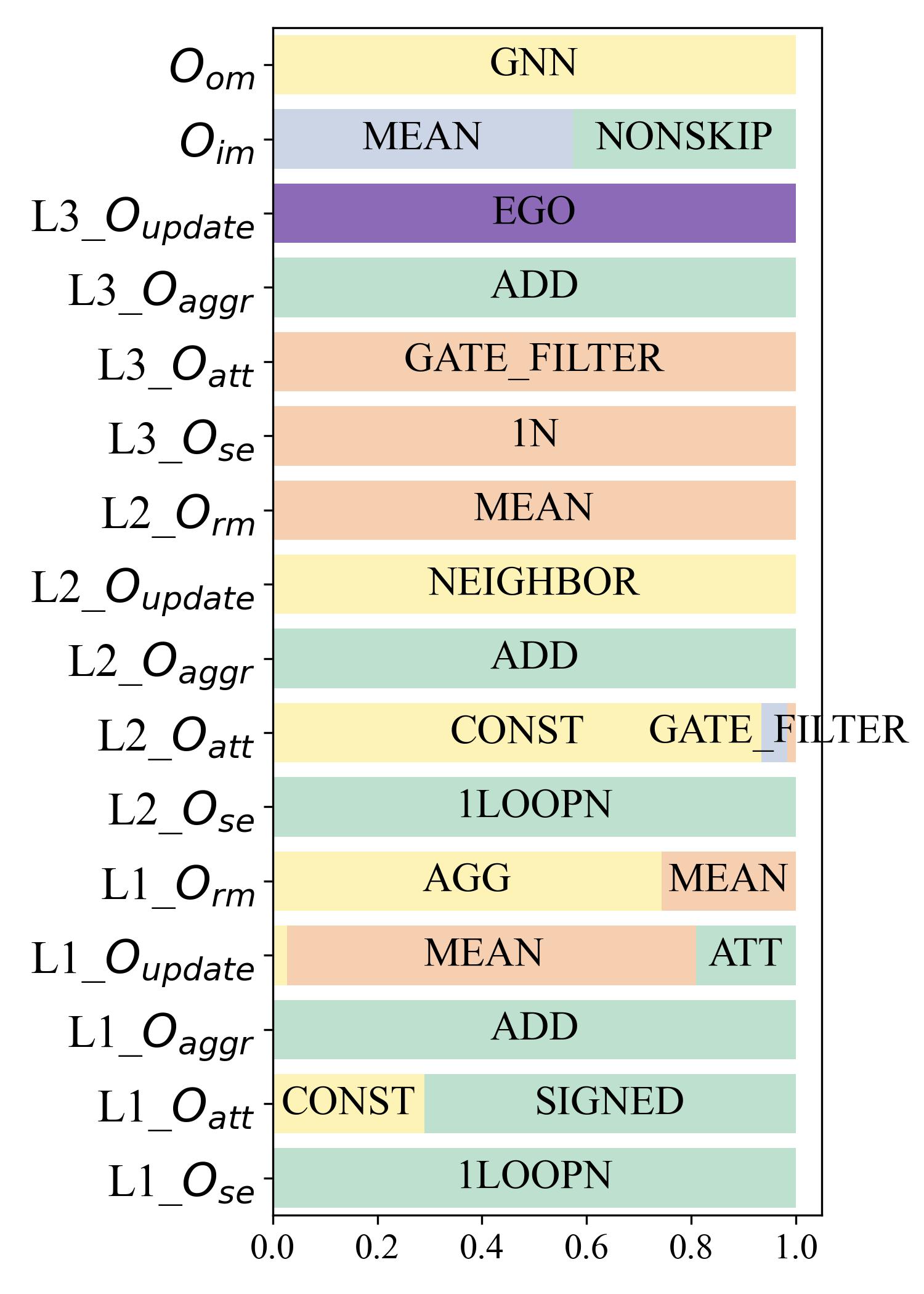}
\end{minipage}
\end{table*}

\subsection{Performance Comparisons}
We show the experimental results of five heterophilous graphs in Table~\ref{tb-performance}. 
As to general GNNs, MLP can achieve better performance than these GNNs on two datasets, and outperformed on the other three datasets. By simply combining the MLP and GCN methods (denoted as MLP+GCN \cite{ma2021homophily} in this paper), it could achieve higher performance and these baselines, which verified the importance of non-aggregated features from the node itself in improving the model performance.
Compared with general baselines, those GNNs designed for heterophilous graphs have higher performance on these datasets, and it demonstrates the effectiveness of non-local neighbors and better GNNs for the heterophilous graphs which have been incorporated in the proposed IIE-GNN.
Designing the GNNs adaptively are helpful in improving the model performance in general, and designing node-wise GNN could further boost the model expressiveness.
For our method, the proposed IIE-GNN can outperform all these baselines on four datasets out of five, and take second place on the left dataset. It owes to incorporating the information from itself and the intra-class neighbors by designing the node-aware GNNs.
We further visualize the operation distributions in each block of the searched node-wise GNNs on the Texas dataset in Table~\ref{tb-performance} (right). It is clear to observe that different nodes have different preferences when utilizing the operations, i.e. different colors exist in the blocks rather than a shared operation like genenral GNNs. It indicates the importance of node-wise GNNs for the heterophilous graphs. 
For the Texas dataset on which the GCN outperformed by MLPs, the \texttt{SIGNED} and \texttt{GATE\_FILTER} operations dominate the Attention blocks. They attach signed values for neighbors when constructing the message, which can enhance the intra-class information when constructing the messages.

\subsection{Evaluations on Intra-class Information Extration}
To evaluate how much intra-class information is utilized in the output vector, we propose a measurement based on the label propagation results.
To be specific, we first preserve the learned weights after training is finished, i.e., the edge weights $e_{uv}$ in each layer and the input feature vectors' weights in the \texttt{ATT} and \texttt{LEARN\_ATT}. Then, we remove all the learnable parameters in the GNN, and aggregate the one-hot labels $\textbf{Y} \in \mathbb{R}^{|\mathcal{V}|\times C}$ along with the designed node-wise GNNs.
Based on the output $\bz \in \mathbb{R}^{|\mathcal{V}|\times C}$, the utilization of intra-information ratio can be calculated as shown in
\begin{align}
	\bz = f_{net}(\textbf{Y}, \bA),
	h_{iir} = \frac{1}{|\mathcal{V}|}\sum_{u \in \mathcal{V}}\frac{z_u^c}{\sum_{j=1}^C z_u^j},
\end{align}
in which $f_{net}$ represents the network mentioned before, $z_u^i$ is the $i$-th column of node $u$'s output, and $c$ is the label of node $u$. 
Based on the intra-class indicator, i.e., one-hot labels, the utilizing and updating procedures on the one-hot labels are equivalent to the intra-class information extraction procedures.
Higher $h_{iir}$ values represented more intra-class information are extracted.
Compared with two measurements mentioned in Section~\ref{sec-preliminary}, $h_{iir}$ provides the dynamic evaluations on the designed graphs, i.e., either by designing the graph structures or updating the edge weights, rather than on the fixed and unweighted structures provided by the input graphs. 
Besides, considering different GNNs have different abilities in utilizing the intra-class information, $h_{iir}$ can take this aspect into consideration which is superior to existing measurements.

\begin{figure}
	\centering
	\includegraphics[width=0.7\linewidth]{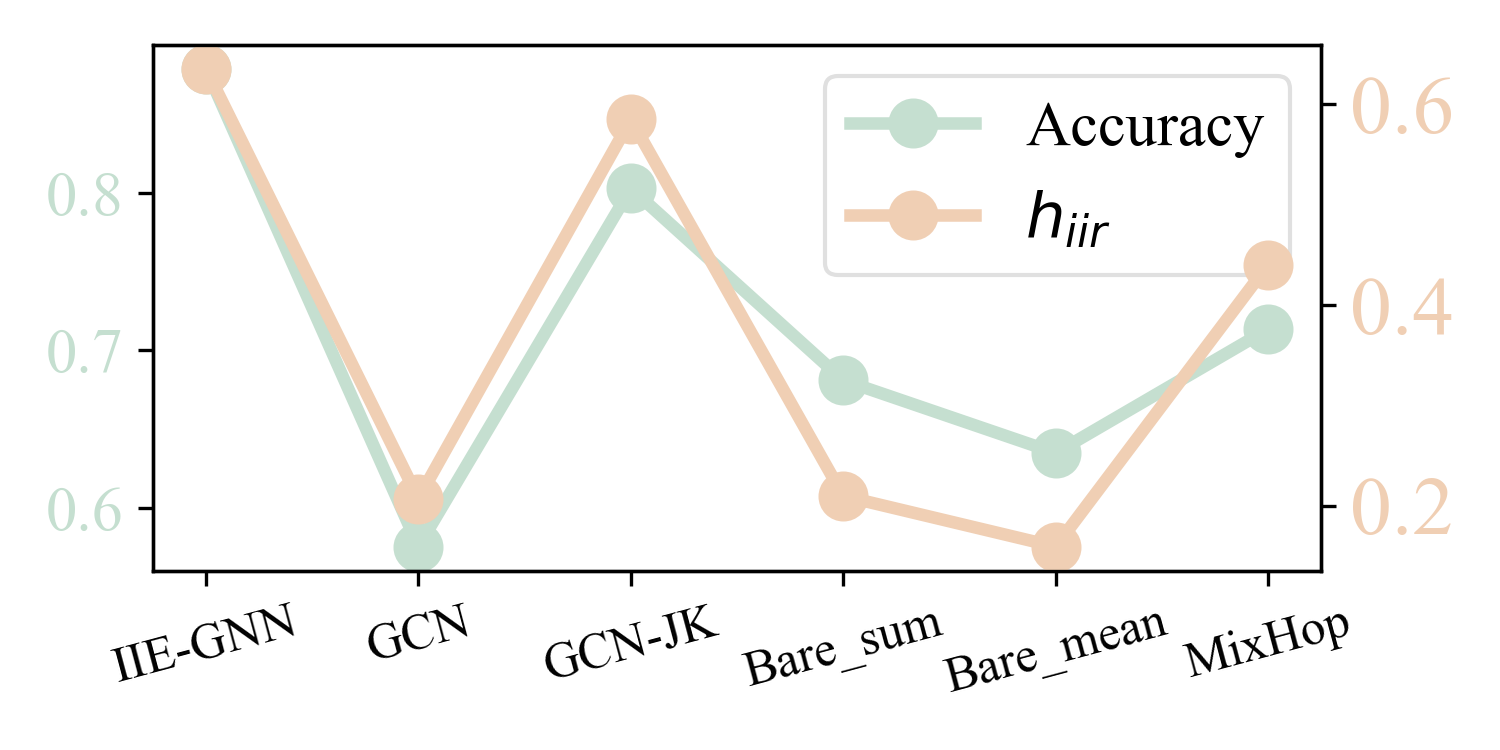}
	\caption{The test accuracy and the intra-class information extraction ratio $h_{iir}$ comparisons on dataset Texas. }
	\label{fig-acc-hiir}
\end{figure}

As shown in Figure~\ref{fig-acc-hiir}, we evaluate the intra-information ratio in different baselines and the proposed IIE-GNN on the Texas dataset. Bare\_mean and Bare\_sum represent the three-layer baselines that use basic aggregation operation with summation and mean when constructing the messages, respectively.
Methods with higher performance have higher $h_{iir}$ values in general. 
Our methods have the highest performance and intra-information ratios than these baselines, which demonstrate the effectiveness in extracting the intra-class information and its effectiveness for the heterophilous graphs.

\section{Conclucion}
In this paper, we propose a novel method IIE-GNN for the heterophilous graphs to enhance the intra-class information extraction. To achieve this, we design a unified framework which contains seven blocks from four levels, on top of which a novel search space is designed. Considering that nodes with different homophily ratios have diverse preferences when designing GNNs, we provide an architecture predictor to design node-wise GNNs based on the search space. We conduct experiments on five heterophilous datasets, and the SOTA performance demonstrates the effectiveness of IIE-GNN. We further evaluate the intra-class information extraction ratios, on top of which the SOTA performance can be analyzed.

\bibliographystyle{ACM-Reference-Format}
\bibliography{main}

\end{document}